\documentclass{optica-article}

\journal{opticajournal} 

\articletype{Research Article}

\usepackage{lineno}

\begin{document}

\title{Sparsity-based background removal for STORM super-resolution images}

\author{Patris Valera,\authormark{1,2,*} Josué Page Vizcaíno,\authormark{1,2} and Tobias Lasser\authormark{1,2}}

\address{\authormark{1}Munich Institute of Biomedical Engineering, Technical University of Munich, Germany\\
\authormark{2}TUM School of Computation, Information and Technology, Technical University of Munich, Germany}

\email{\authormark{*}patris.valera@tum.de} 


\begin{abstract*} 
Single-molecule localization microscopy techniques, like stochastic optical reconstruction microscopy (STORM), visualize biological specimens by stochastically exciting sparse blinking emitters. The raw images suffer from unwanted background fluorescence, which must be removed to achieve super-resolution. We introduce a sparsity-based background removal method by adapting a neural network (SLNet) from a different microscopy domain. The SLNet computes a low-rank representation of the images, and then, by subtracting it from the raw images, the sparse component is computed, representing the frames without the background. We compared our approach with widely used background removal methods, such as the median background removal or the rolling ball algorithm, on two commonly used STORM datasets, one glial cell, and one microtubule dataset. The SLNet delivers STORM frames with less background, leading to higher emitters' localization precision and higher-resolution reconstructed images than commonly used methods. Notably, the SLNet is lightweight and easily trainable (<5  min). Since it is trained in an unsupervised manner, no prior information is required and can be applied to any STORM dataset. We uploaded a pre-trained SLNet to the Bioimage model zoo, easily accessible through ImageJ. Our results show that our sparse decomposition method could be an essential and efficient STORM pre-processing tool.
\end{abstract*}

\section{Introduction}
Super-resolution microscopy (SRM) assembles a group of optical microscopy techniques that allow fluorescent imaging at (sub-)cellular level and at unprecedented resolutions. Unlike the traditional light microscopy techniques, these methods exceed the Abbe diffraction resolution limit, resulting in a 10-nanometer distance limit between two light point sources and a lateral resolution of 20 nanometers \cite{Galbraith:01, Schermelleh:02}. One type of SRM is single-molecule localization microscopy (SMLM) \cite{Galbraith:01, Schermelleh:02, Huang:03, Lelek:04, Klein:05}, which uses the activation of fluorophores to visualize biological structures and is categorized into (fluorescence) photo-activation localization microscopy ((f)PALM) \cite{Hess:06, Sengupta:07} and (direct) stochastic optical reconstruction microscopy ((d)STORM) \cite{Rust:08, Xu:09}. SMLM methods stochastically activate a sparse set of fluorophores at different time steps and capture these in multiple image acquisitions. In a post-processing step, the centroids of the emitters are estimated and accumulated into a high-resolution reconstructed image with nanometer-level precision \cite{Galbraith:01}. 

\begin{figure}[ht!]
    \centering\includegraphics[width=13cm]{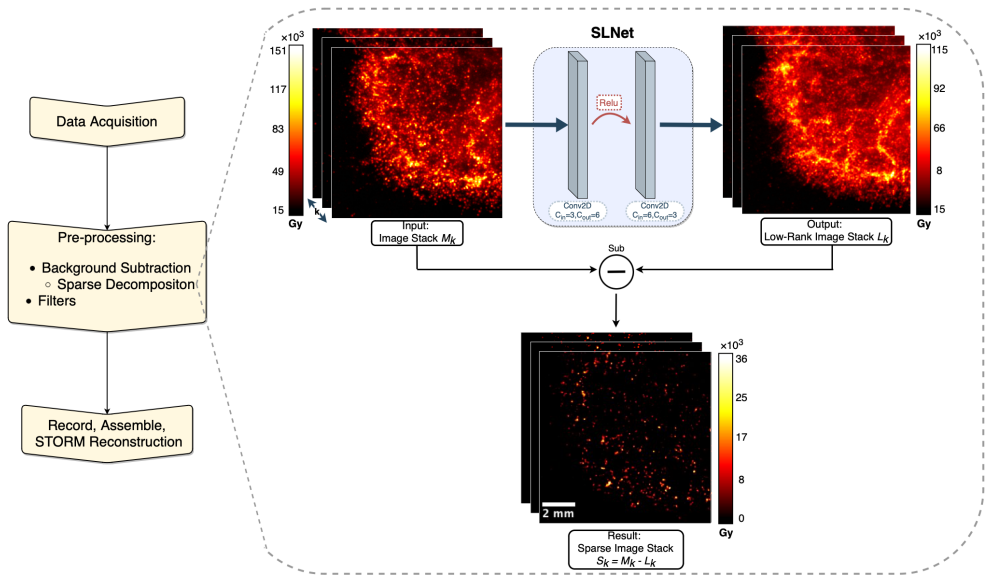}
    \caption{\textbf{Left:} The three main steps conducted in STORM, where the background removal methods are applied in the pre-processing stage. \textbf{Right:} The sparse decomposition process and the architecture of the neural network SLNet. The network approximates the low-rank representation stack from the input stack, and the sparse images are calculated as a subtraction of the two. The parameter  \emph{k} controls the size of the input stack (here, \(k=3\)).}
    \label{fig:sparse_decomposition_slnet}
\end{figure}

One drawback of SMLM is its sensibility to background signals often present in fluorescence microscopy (e.g., autofluorescence and non-bleached out-of-focus molecules \cite{Xu:09, Van:10}). 
These signals affect the ability to accurately localize the emitters' centroids within the sample, known as localization precision \cite{Baddeley:11, Waters:12}, ultimately leading to decreased image resolution and artifacts in the reconstruction. Therefore, a background removal pre-processing (see Fig. \ref{fig:sparse_decomposition_slnet}) is crucial for precise and high-resolution image reconstruction in STORM. 

There have been several attempts to remove the background, such as median background removal \cite{Ambrose:13, Huang:14}, "rolling ball" algorithm \cite{Sternberg:15,Rashed:16}, or image filtering \cite{Deschout:17, Chandel:18}, but most are either too general or too computationally expensive. The background in fluorescence microscopy is heterogeneous and non-uniform, but most methods assume otherwise \cite{Deschout:17}. 
For example, a standard way is to compute the median intensity of all frames and subtract it from them to get images without a prevalent background. Although it may be simple, this method generally removes an atypical uniform background that leaves unnecessary fluorescence in the images or can also wrongly remove fluorophores. On the other hand, the "rolling ball" algorithm removes the background by "rolling" a ball of a certain radius under the pixel intensity histogram of each frame and then subtracts the area below it. This algorithm is an intuitive method to remove uniform background intensity levels, but as such, it has limitations; for example, at times, it tends to include a disproportionately large amount of weak spots or generate more artifacts in the background \cite{Sternberg:15}. Another widely used technique is image filtering, which includes temporal and band-pass filters applied to correct the background. However, they have proven insufficient in cases with structured and non-uniform backgrounds \cite{Deschout:17, Mockl:19}. 

As a novel and refined solution to the aforementioned problems, we propose to use the SLNet \cite{Vizcaino:20}, previously used for background removal in zebrafish light field 3D microscopy. The SLNet is a fast, computationally inexpensive, sparse-low-rank decomposition-based method to compute and remove the background of fluorescent static images (see Fig. \ref{fig:sparse_decomposition_slnet}). 
It is trained in an unsupervised fashion by minimizing a custom-crafted loss function that aims to approximate the raw images with a low-rank representation using three sample images at different time points. Then, the sparse components, the images without the background, are computed by subtracting the low-rank components from the raw images. 

This work compares the SLNet and different background removal methods as a pre-processing step, aided by ThunderSTORM \cite{Ovesny:21}, an open-source plug-in for ImageJ \cite{Schneider:22}, which performs localization and reconstruction. A hyper-parameter optimization of the SLNet was also performed, including parameters like the number of images, learning rate, the required training epochs, etc.

In \autoref{ssec:SLR_decom}, we first present the sparse-low-rank decomposition on which our method is based. Then, in \autoref{ssec:SLNet}, we describe the SLNet neural network adapted from \cite{Vizcaino:20} and explain the unsupervised training of the SLNet in \autoref{ssec:SLNet_training}. The datasets we used are defined in \autoref{ssec:datasets}, and the reconstruction method of the super-resolution images is presented in \autoref{ssec:reconstruction}. In the experimental results in \autoref{sec:experiments}, we describe the methods we used to compare our findings and present our quantitative evaluation in \autoref{ssec:quantitative} with 2 FWHM analyses in \autoref{ssec:fwhm} and a SQUIRREL analysis in \autoref{ssec:SQUIRREL}. In the hyperparameter search results in \autoref{ssec:alpha_mu}, we describe the effect of 2 crucial parameters on the SLNet training and the sparsity output level.

\section{Methods}\label{sec:methods}
In short, our method uses the sparse decomposition implemented with the help of the neural network SLNet by assuming that the blinking emitters' raw frames used for the STORM reconstruction consist of low-rank components representing the background and sparse components representing the specimen's essential signals.

\subsection{Sparse-low-rank Decomposition}\label{ssec:SLR_decom}
The sparse-low-rank decomposition (SD) method, also known as robust principal component analysis (PCA) \cite{Lin:23, Yuan:24}, computes the low-rank and sparse components of a matrix $M$ with the constraint that the sum of the components equals the matrix $M$. In our case, the data is a set of images with size \(m \times n\) and recorded at \(k\) time points, giving us \(M_{k,m,n} \in \mathbb{R}_{\geq 0}^{k \times m \times n}\). We can use sparse decomposition to decompose temporal stacks of the frames into sparse and low-rank components by rearranging \(M_{k,m,n}\) as \(M_{k,mn} \in \mathbb{R}_{\geq 0}^{k \times mn}\) and minimizing the convex optimization problem: 
\begin{equation}
\label{eqn:optimization_problem}
\begin{split}
\min_{L, S} & \quad {\lVert L \rVert}_* + \lambda {\lVert S \rVert}_1  \\
\text{subject \ to} & \quad L + S = M_{k,mn}
\end{split}
\end{equation}
where
\begin{itemize}
  \item \(L \in \mathbb{R}_{\geq 0}^{k \times mn}\) is the low-rank component
  \item \({\lVert L \rVert}_*\) denotes the nuclear norm (the sum of all singular values) of the low-rank component
  \item \(S \in \mathbb{R}_{\geq 0}^{k \times mn}\) is the sparse component
  \item  \({\lVert S \rVert}_1\) denotes the \(L_1\)-norm (component-wise sum of absolute values of all entries) of the sparse component
  \item \(\lambda > 0\) is the weighting parameter that controls the ratio between the low-rank and sparse components and thus the level of sparseness.
\end{itemize}

This convex optimization problem is NP-hard, as it requires minimizing both \(L_1\)-norm and nuclear norm at the same time. Different solutions have been found to solve this problem, such as the iterative thresholding approach \cite{Beck:25, Wright:26}, the accelerated proximal gradient approach \cite{Toh:27}, the dual approach \cite{Lin:28}, and methods of augmented Lagrange multipliers \cite{Lin:23}, which are primarily used due to faster convergence \cite{Lin:23, Yuan:24}. However, we solve it with the procedure proven by \cite{Herrera:29} and used by \cite{Vizcaino:20}, where the low-rank matrix \(L\) is computed with a neural network. Then, the sparse matrix \(S\) is calculated as the subtraction of the original image with the low-rank component, equaling \((M-L)_{\geq 0}\).

\subsection{Neural Network SLNet}\label{ssec:SLNet}
The SLNet comes into action for approximating the low-rank component of the input:
\begin{equation}
\label{eqn:slnet}
\mathcal{N}_{\Theta}^{SL}(M_{k,m,n}) \approx L_{k,mn}
\end{equation}
It takes as an input the set of images \(M_{k,m,n}\) and generates the low-rank component \(L_{k,mn}\) of the stack of images, parametrized by \(\Theta\). After finding the low-rank component, the sparse component S (the resulting images) is calculated as 
\begin{equation}
S = (M_{k,mn} - \mathcal{N}_{\Theta}^{SL}(M_{k,m,n}) )_{\geq 0}
\end{equation}

SLNet is a simple and small neural network consisting of two convolutional layers and one rectified linear unit (ReLU) activation function (see Fig. \ref{fig:sparse_decomposition_slnet}). This small architecture brings the benefits that the network takes less space and performs an efficient computation. The network takes a stack of raw images \(M_k\) converted to a tensor with the shape \((batch\_size, k, m, n)\) as input and outputs a tensor with the same size, which represents the low-rank components of the input images. The weight initialization of SLNet is essential for the output \(L\). If the weights are too large, the output \(L\) has higher entries than the input, and the resulting sparse component \(S\) from their subtraction would have only \(0\)s as its elements, and thus the learning would stop. Therefore, the weights \(\Theta\) are first initialized with the Kaiming method \cite{He:30}, which considers the non-linearity of the ReLU activation function and prevents the exponential growth or shrinking of the input. 
For further details on the architecture and training methodology, we refer the reader to the publicly available source code \cite{Valera:31}.

\subsubsection{Network Training}\label{ssec:SLNet_training}
The network is trained in an unsupervised manner by minimizing the augmented Lagrangian loss function:
\begin{equation}
\label{eqn:loss_function}
\min_{\Theta} \quad {\lVert M - \Gamma_\mu(\mathcal{N}_{\Theta}^{SL}(M_{k,m,n})) \rVert}_1 + \alpha \cdot \overline{\lvert S \rvert} + \overline{\lvert R \rvert}
\end{equation}
with \(M\) the raw images, \(\alpha \cdot \overline{\lvert S \rvert}\) the scaled average of the sparse component \(S\) weighted by an \(\alpha\) parameter, and \(\overline{\lvert R \rvert}\) the average reconstruction error.
The operator \(\Gamma_\mu(\cdot)\) performs singular value shrinkage \cite{Leeb:32, Barash:33} and enforces its output to have a low rank by setting the eigenvalues of the rectangular matrix of the input's SVD to 0 if they are smaller than a threshold \(\mu\), meaning \(\Sigma_{<\mu}=0\) and decreasing the remaining eigenvalues. Mathematically, the operator is expressed as:
\begin{equation}
\label{eqn:shrinkage}
\begin{split}
\Gamma_\mu(X) =\ & U\ [\mathrm{sign}(\Sigma) \cdot \max(|\Sigma| - \mu, 0)]\ V^*  \\
& \text{where } X = U\Sigma V^*
\end{split}
\end{equation}

The threshold \(\mu\) and the parameter \(\alpha\) are critical user-defined parameters that control the degree of the low rank of the SLNet output \(L\) and the level of sparseness of the result \(S\). The network training was configured with a sparseness threshold \(\mu\) set to 0.01 and the alpha parameter set to 12. This configuration was selected based on the sparsity of the output frames from differently trained neural networks and the reconstruction quality they provided. The effects of these parameters are further explained in \autoref{ssec:alpha_mu}. 

We trained the network for 100 epochs because it produced satisfactory sparse results and reconstructions. The training took 2 minutes and 37 seconds on a workstation with an AMD Ryzen 9 3950X 16-core Processor, and using an NVIDIA GeForce RTX 2080 Ti GPU. The training and testing took about 8 minutes. For the sparse decomposition, we use only three frames at a time, chosen based on a time index \(t\geq 0\) that shifts them. The frames are \(M_t\), \(M_{t+50}\) and \(M_{t+100}\) (derived from the original SLNet \cite{Vizcaino:20}). Consequently, the trained SLNet can be used in parallel after the microscope records 100 specimen images. 

\subsection{Dataset descriptions}\label{ssec:datasets}
We used two datasets with distinct samples to train and test the neural network and generate other methods' results. The first experimental data we used is the dataset obtained from \cite{Leterrier:34}, which includes actin-labeled glial cells of a rat's hippocampal neurons. We trained the network with the first 100 images of the BIN10 frames and tested on the remaining 5890 frames. The second is a microtubule dataset obtained from \cite{Kumar:35}, consisting of microtubules immunolabeled with Alexa Fluor 647 phalloidin. We used the first 7000 out of 60000 frames for testing. 


\subsection{STORM super-resolution reconstruction}\label{ssec:reconstruction}
After acquiring the images and pre-processing them to get high-quality raw images, the most critical step is the STORM high-resolution image reconstruction performed using the ImageJ \cite{Schneider:22} plugin ThunderSTORM \cite{Ovesny:21}. To compare the different results and find out the most suitable combination of image filters \cite{Ovesny:36} with background removal methods, the image pre-processing was done with all the supported filters on ThunderSTORM:
\begin{itemize}
    \item averaging filter (kernel size 3 pixels),
    \item Gaussian filter (sigma 1.6 pixels),
    \item lowered Gaussian filter (sigma 1.6 pixels),
    \item difference-of-Gaussians filter (sigma1 1.0 pixels, sigma2 1.6 pixels),
    \item difference of averaging filters (first kernel size 3 pixels, second kernel size 5 pixels),
    \item wavelet filter (B-spine, order 3, scale 2),
    \item median filter (box pattern with kernel size 3 pixels).
\end{itemize}
Additionally, the first testing dataset (the glial cells) contains a high density of emitters, and as a fluorescence live cell imaging, the fluorophores of the neurons are on the move. Consequently, the centroids of the fluorescent molecules are distant, and we needed to post-process the reconstructed images with lateral drift correction with cross-correlation (five bins at 5 x magnification). 

\section{Experiments and Results} \label{sec:experiments}
To assess our method, we compared the following pre-processing methods:
\begin{itemize}
    \item The raw non-preprocessed images
    \item Median background removal (implemented in Python by subtracting the median value of all pixels in all frames to the raw data)
    \item Rolling ball background subtraction (provided as the plug-in "Rolling Ball Background Subtraction" \cite{rolling_ball_plugin:37, rolling_ball_plugin2:38} in ImageJ. The glial cell and microtubules datasets were processed without smoothing and a radius of 3 and 10 pixels, chosen  heuristically.)
    \item The SLNet with $\alpha=12$ and $\mu=0.01$.
\end{itemize}

\begin{figure}[ht!]
\centering\includegraphics[width=\textwidth]{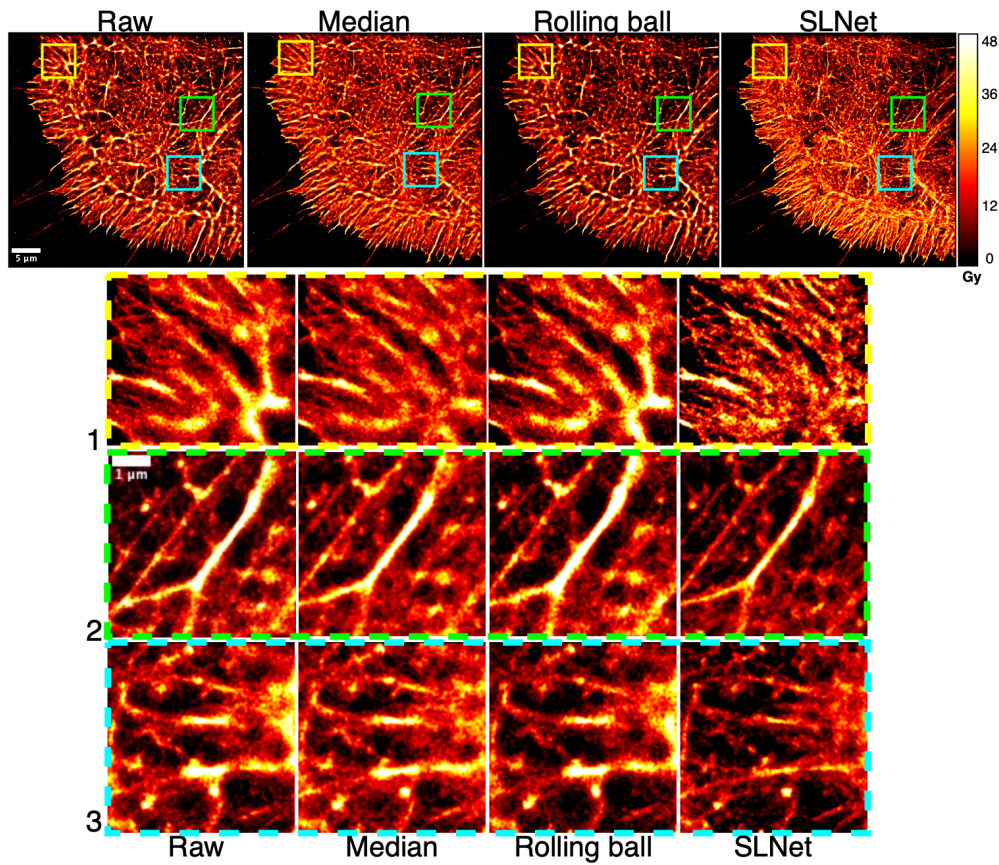}
\caption{Super-resolution reconstruction of glial cells using ThunderSTORM and three background removal methods: median, "rolling ball" with radius 3 pixels and our SLNet background removal method. (1), (2), (3) Magnified views of three selected regions with scale bar $1\mu m$.}
\label{fig:zerocost_fwhm}
\end{figure}

\begin{figure}[ht!]
\centering\includegraphics[width=\textwidth]{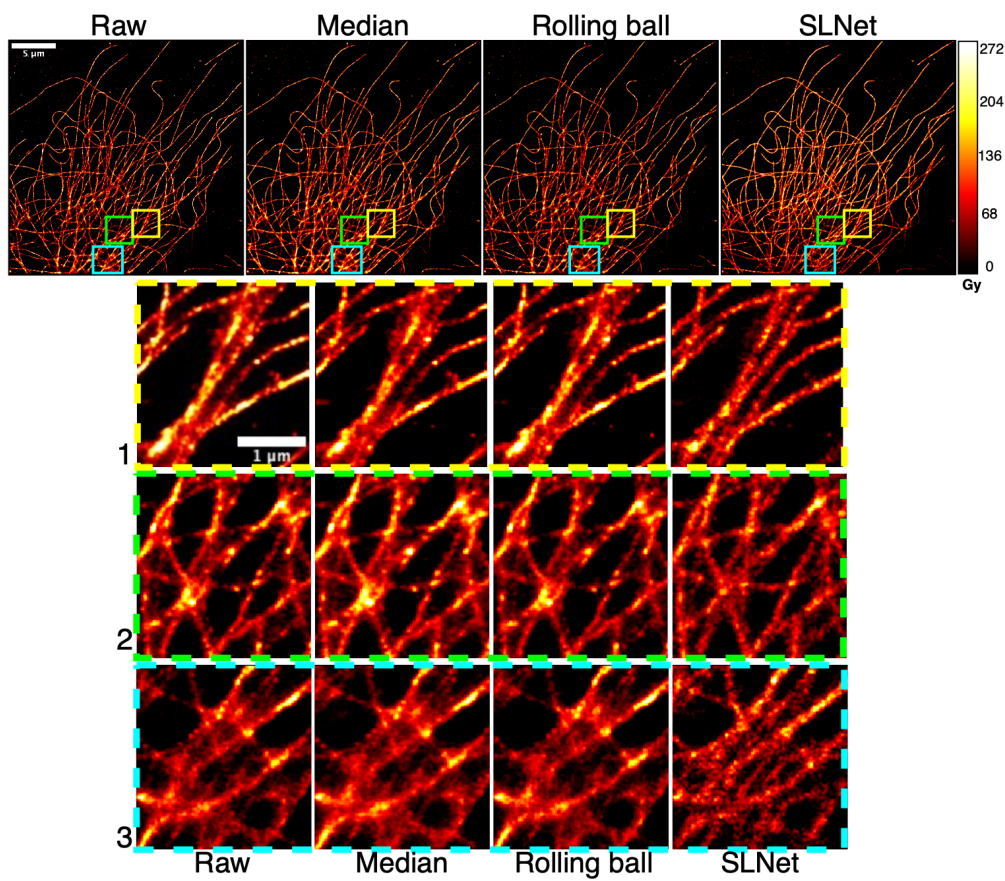}
\caption{Super-resolution reconstruction of microtubules using ThunderSTORM and three background removal methods: median, "rolling ball" with radius 10 pixels and our SLNet background removal method. (1), (2), (3) Magnified views of three selected regions with scale bar $1\mu m$.}
\label{fig:microtubules_fwhm}
\end{figure}

These methods are compared on the STORM datasets (described in sec.~\ref{ssec:datasets}) qualitatively in figures~\ref{fig:zerocost_fwhm} and \ref{fig:microtubules_fwhm}, and quantitatively using the metrics described in sec.~\ref{ssec:quantitative}, presented in figures~\ref{fig:box_plot} and \ref{fig:nanoj}. 

The reconstruction of the raw images was challenging due to high autofluorescence caused by the densely packed fluorophores. Figures \ref{fig:zerocost_fwhm} and \ref{fig:microtubules_fwhm} show the best reconstruction of each method and three magnified areas with densely packed structures to illustrate the improvement of our method. The SLNet outputs have individual emitters sufficiently separated, leading to a better approximation of the coordinates of centroids and resulting in sharper and more defined images compared to other methods.


\subsection{Quantitative comparison} \label{ssec:quantitative}
As super-resolution methods achieve sub-Abbe-limit resolutions, no methodology exists to generate or measure ground-truth data. Hence, data-driven approaches are used to compare different pre-processing and reconstruction methods, as described in this section.

\subsubsection{FWHM evaluation}\label{ssec:fwhm}
\begin{figure}[ht!]
\centering\includegraphics[width=7cm]{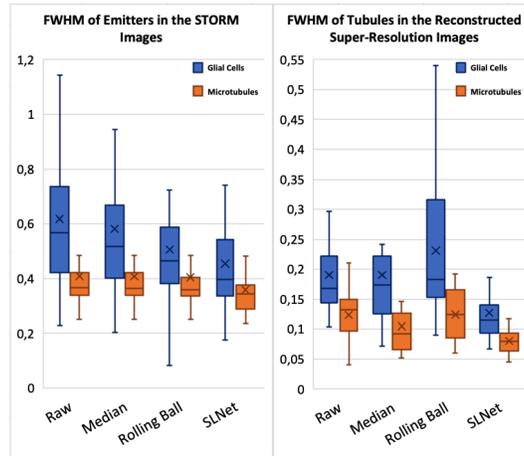}
\caption{The measurements were taken for the raw frames and the three methods (SLNet, Median Background Removal, and "Rolling Ball" algorithm) on the glial cells and microtubules dataset \cite{Leterrier:34, Kumar:35}. \textbf{Left:} Box plot presenting quantitative results of around 45 measured emitters' FWHM on the STORM emitter images. \textbf{Right:} Box plot presenting quantitative results of around 20 measured tubules' FWHM of the reconstructed super-resolution images.}
\label{fig:box_plot}
\end{figure}

To evaluate the resolving power of the STORM reconstructions, we analyzed the full width at half maximum (FWHM) of:
\begin{itemize}
    \item The single emitters after pre-processing, where we extracted the position of emitters (38 for the glial cells and 45 for the microtubules) on a single thresholded frame of the raw data using the "Analysing Particles..." \cite{analyzing_particles_command:39} command of ImageJ.
    \item Single microtubules/filament extensions on the reconstructed images from both datasets. We used 20 microtubules to measure their transversal FWHM and evaluate the statistical distribution. We manually selected microtubules that were visually small and close to each other, thus challenging to reconstruct due to the proximity of the fluorophores in those areas.   
\end{itemize}
Fig.~\ref{fig:box_plot} shows the FWHM analysis on both datasets for the emitter and the reconstructed images. 

The SLNet FWHM measurements show a better central tendency for both datasets since the median is lower than the other methods in all cases. As a result, our method appears better at improving spatial resolution. 
We observed that the other methods had some extreme outliers, whereas in the SLNet results, the measurements were more uniform. In addition, our method has shown symmetry in all measurements (especially in the FWHM analysis of the reconstructed glial cells), generally leading to a bias towards lower FWHM values. From these results, we concluded that the sparse decomposition tends to produce lower FWHM values, has less variability, and is less prone to outliers.

\begin{figure}[ht!]
\centering\includegraphics[width=13cm]{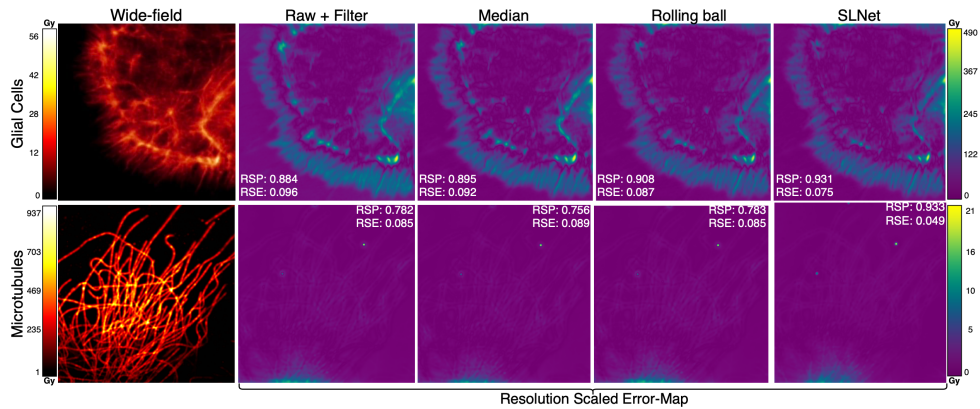}
\caption{Wide-field images of both datasets and SQUIRREL analysis comparing the four respective STORM reconstructions. SLNet delivers better linearity in reconstruction concerning the wide-field image of datasets.}
\label{fig:nanoj}
\end{figure}

\subsubsection{SQUIRREL Analysis}\label{ssec:SQUIRREL}
As a second step of the evaluation, we performed a SQUIRREL analysis (see Fig. \ref{fig:nanoj}) with the help of NanoJ- SQUIRREL \cite{Culley:40}, an ImageJ plug-in. SQUIRREL stands for Super-resolution QUantitative Image Rating and Reporting of Error Locations, because it assesses and maps errors and artifacts of super-resolution images concerning their dataset wide-field image. A SQUIRREL analysis calculates the scaled Pearson's coefficient (RSP) and the smallest resolution scaled error (RSE). A higher RSP and a lower RSE indicate a better agreement and linearity of the reconstruction with the wide-field image. Our conclusions based on the FWHM analysis were confirmed by analyzing each method's best-reconstructed images with NanoJ-SQUIRREL. It proved that the SLNet super-resolution STORM image had the highest RSP values and the smallest RSE for both datasets, hence having the highest agreement with the wide-field image from all other STORM frames. The "rolling ball" images achieved the second-best agreement.

\subsection{SLNet hyperparameter search: Sparsity threshold and alpha}\label{ssec:alpha_mu}
The threshold \(\mu\) from the SLNet singular value shrinking operator (\autoref{eqn:shrinkage}) and the \(\alpha\) parameter from the loss function (\autoref{eqn:loss_function}) control the low rank of the SLNet predicted dense components and, consequently, also the sparsity of the output sparse frames. The \(\mu\) parameter indicates how the eigenvalues of the input frames should be shrunk to lower the input frames' rank and to produce the sparse component at the end. To demonstrate this behavior affected by the \(\mu\) threshold, we trained the SLNet for 100 epochs on 100 images with different \(\mu\) parameters, i.e., with \(\mu = \{0,\ 0.2,\ 0.5,\ 1,\ 3,\ 10,\ 20,\ 30\}\). On the other hand, the \(\alpha\) parameter influences the loss function by weighting the sparse component. We trained the network with \(\alpha = \{0,\ 0.2,\ 0.5,\ 1,\ 3,\ 7,\ 12,\ 20,\ 30\}\) and tested with the remaining raw frames. Both parameters are user-defined and influence our background removal method directly. 

\begin{figure}[ht!]
\centering\includegraphics[width=13cm]{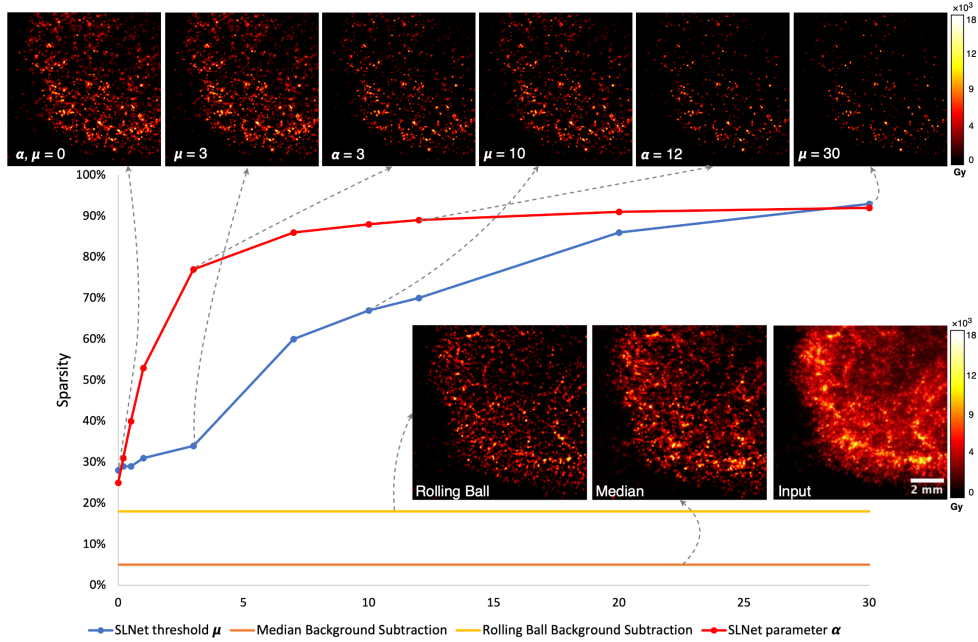}
\caption{Comparison of the sparsity level of glial cells \cite{Leterrier:34} frames from SLNets trained with various \(\mu\) and \(\alpha\) values with the median background removal and the "rolling ball" algorithm images.}
\label{fig:sparsity_graph}
\end{figure}

The degree of sparseness (sparsity) is the percentage of zero elements out of all the elements as in: 
\begin{equation}
\label{eqn:sparsity}
sparsity = \frac{\#\ zero\ elements}{\#\ total\ elements} \times 100
\end{equation}
Fig. \ref{fig:sparsity_graph} shows how these parameters influence the level of sparseness compared with the other methods. The results verify the idea behind the sparse decomposition and our unsupervised training of the network with the help of the loss function because the sparsity of the results increases as the threshold \(\mu\) and the \(\alpha\) parameter increase. As seen, even when both parameters are set to 0, the network produces a low-rank starting point, and the decomposition outputs a sparser result. The \(\alpha\) parameter influences the sparsity faster than the \(\mu\) threshold since, as seen from the graphs, for \(\alpha\) = 3, we get 77\% sparsity, whereas for \(\mu\) = 3, 34\%. 

Additionally, all SLNet results have a higher sparseness level than the other method's outcomes, constant at either 5\% or 18\%. As such, our method implies numerous benefits, such as significantly reducing the computational time, the duration of the STORM reconstruction, and the flexibility to alter the background levels based on the concrete data. By analyzing these results and considering the varying background levels in each dataset, we concluded that both parameters are crucial, and the user should choose them based on the dataset and the desired sparseness level. Nevertheless, the user should consider that training the network for too long or choosing too high values for both parameters, the output images would have very high sparsity and be counterproductive for the STORM reconstruction. 

\section{Discussion}
On the grounds of implementing a more performant approach to deal with the problem of constant or autofluorescence background that occurs in high-resolution fluorescence microscopy, we introduce and adapt a sparse decomposition-based background removal method using the neural network SLNet (fig. \ref{fig:sparse_decomposition_slnet}) in working with STORM raw images. Subsequently, SLNet proved to be an efficient option as a convolutional neural network to produce the low-rank components of the input images and perform the sparse decomposition. This decomposition computes the sparse images without the unnecessary background, ready to be used for the STORM super-resolution reconstruction. 

The sparsity-based background removal method presented is computationally efficient due to the simple architecture of the SLNet and the fast training and evaluation time. Furthermore, we demonstrated how the parameters \(\mu\) and \(\alpha\) control the sparseness level of the output frames due to their influence on the loss function (fig. \ref{fig:sparsity_graph}). Both these parameters are very user-friendly because they are easily reconfigured for the model training. They are also crucial in STORM since the studied samples and used microscopic systems differ in the level of fluorescence they produce. Hence, the user can control the sparseness based on the present background and obtain satisfactory decomposition results. 

To evaluate our method, we compared the reconstruction of other commonly used methods, such as the median background removal and the "rolling ball" algorithm (fig. \ref{fig:zerocost_fwhm}, \ref{fig:microtubules_fwhm}). The reconstruction was executed with the software ThunderSTORM, and we encountered that the most satisfactory results were derived from a combination of image filters (part of the ThunderSTORM workflow) and the SLNet. Additionally, the reconstruction of our outcomes was faster (up to 80\% faster for a reconstruction without any filters) than the other methods due to the sparsity of the frames. Measuring the FWHM of the emitters in the STORM frames and small areas like microtubules or thin filament extensions in the reconstructed images (fig. \ref{fig:box_plot}) delivered lower FWHM values and proved that our sparse decomposition method achieved the highest resolution. A SQUIRREL analysis (fig. \ref{fig:nanoj}) proved the better agreement and linearity of the SLNet's result out of all other super-resolution reconstructed images compared with the respective wide-field frame. 

Overall, our method accurately processed sparse frames with less background, leading to a high localization precision of the emitters' centroids. Consequently, the fluorophores were adequately distant and visible, and the specimen's structure was defined and shaper. In contrast, the other reconstructed images were blurry in highly dense areas, and the tubules appeared fused with each other. However, we saw (Fig. \ref{fig:microtubules_fwhm} bottom right tubules) that the sparse decomposition is not beneficial when parts of the specimen occur in distant image regions since it approximates those molecules as background and removes them.  

\section{Conclusion}
In this paper, we presented a method to remove the background of STORM images using a sparse decomposition that assumes the background represents the low-rank components and the specimen represents the sparse ones. It was implemented by adapting the neural network SLNet to work on STORM, and we assessed that it outperformed some of the most used approaches by scientists. The best case to use SLNet is when studying specimens with a detailed and overlapping structure. Our method could be a beneficial tool in the STORM microscopy environment that is specific to its input and delivers high-resolution images due to its fast computational time and the flexibility offered in the image sparseness levels.

\begin{backmatter}

\bmsection{Funding}
Content in the funding section will be generated entirely from details submitted to Prism. Authors may add placeholder text in the manuscript to assess length, but any text added to this section will be replaced during production and will display official funder names along with any grant numbers provided. If additional details about a funder are required, they may be added to the Acknowledgments, even if this duplicates information in the funding section. See the example below in Acknowledgements. For preprint submissions, please include funder names and grant numbers in the manuscript.


\bmsection{Disclosures}
The authors declare no conflicts of interest.

\bmsection{Data Availability Statement} Data underlying the results presented in this paper are available in \cite{Leterrier:34, Kumar:35}. The code is available in \cite{Valera:31}.

\end{backmatter}

\bibliography{bibliography}

\end{document}